\documentclass{svproc}

\usepackage{upmath}
\usepackage[T1]{fontenc}
\usepackage{lmodern}
\usepackage[utf8]{inputenc}
\usepackage[table]{xcolor}
\usepackage{extarrows}
\usepackage{hyperref}
\usepackage{url}     
\usepackage{array, booktabs}
\usepackage{amsfonts}    
\usepackage{nicefrac}   
\usepackage{microtype} 
\usepackage{lipsum}
\usepackage{graphicx}
\usepackage{lscape}
\usepackage{afterpage}
\graphicspath{ {images/} }
\usepackage{ulem}
\usepackage{multirow}
\usepackage{amsmath}
\usepackage{enumitem}
\usepackage{subcaption}
\usepackage{caption}

\usepackage{url}

\newcolumntype{C}[1]{>{\centering\arraybackslash}m{3em}}

\newcounter{bar}
\newcommand{\ex}{%
	 {Example \stepcounter{bar}\thebar: }}

\begin{document}
\mainmatter              
\title{Impact of News on the Commodity Market: Dataset and Results}
\titlerunning{Impact of News on the Commodity Market: Dataset and Results}  
%

\author{Ankur Sinha, Tanmay Khandait}
\institute{
Production and Quantitative Methods\\
Indian Institute of Management Ahmedabad\\
Ahmedabad, India 380015\\
\texttt{\href{asinha@iima.ac.in}{asinha@iima.ac.in}, \href{tanmayk@iima.ac.in}{tanmayk@iima.ac.in}}}

%

\maketitle              

\begin{abstract}
Over the last few years, machine learning based methods have been applied to extract information from news flow in the financial domain. However, this information has mostly been in the form of the financial sentiments contained in the news headlines, primarily for the stock prices. In our current work, we propose that various other dimensions of information can be extracted from news headlines, which will be of interest to investors, policy-makers and other practitioners. We propose a framework that extracts information such as past movements and expected directionality in prices, asset comparison and other general information that the news is referring to. We apply this framework to the commodity ``Gold'' and train the machine learning models using a dataset of 11,412 human-annotated news headlines (released with this study), collected from the period 2000-2019. We experiment to validate the causal effect of news flow on gold prices and observe that the information produced from our framework significantly impacts the future gold price. 
\end{abstract}
\section{Introduction}
News has always been one of the primary sources of influence while making financial decisions. With technology emerging, we have news flow across all domains in amounts that exceed the cognitive capacity of an individual, especially in the field of finance. Recent developments in text mining have enabled investors and policy-makers to capitalize on the unstructured information, primarily textual data, for making decisions.
\par The Semi-Strong Efficient Market Hypothesis \cite{Malkiel1989}, has inspired much research on establishing the relationship between the opinions of widely available information and returns of stock and commodity prices \cite{tetlock2007giving,hess2008commodity}. News, in the form of headlines \cite{ederington1993markets,Automated_Text_Analytics}, opinions \cite{tetlock2007giving} and various surveys have been known to influence the decision making process of the investors significantly. However, the effect of news items in the context of commodities has not received as wide attention as stocks \cite{sinha2019buy,takala2014gold,malo2014good,malo2013learning}. It is known that the prices of the commodities are highly volatile, but very few papers have studied the impact of news on commodity prices, which has the potential to explain some of the uncertainties. In this paper, we bridge this gap by applying news analytics to one of the most important commodities, which is gold. The aim of this study is to provide the investors and policy-makers with a tool to process and analyze large amount of gold-related news that is flowing continuously from multiple geographies. 
\par The works related to public opinions impacting gold commodity prices have emerged in two forms. The first category of papers investigated the effect of various macroeconomic announcements on the price of commodities. Frankel and Hardouvelis \cite{frankel1985commodity} were one of the first to study the effect of the announcements on various commodities. Barnhart \cite{barnhart1989effects} worked on the effect of money supply announcements and macroeconomic indicators on the commodity prices. Macroeconomic announcements were used to study the effect of these on intra-day gold and silver futures \cite{christie2000macroeconomics} and the volatility of gold market \cite{cai2001moves}. Some other work that was built up around similar ideas are \cite{caporale2017macro,elder2012impact,roache2009effects}.
\par The second category of papers investigated the effect of various news from publishing houses, and opinions and sentiments of people from micro-blogging sites, like twitter, on the price of the gold commodity. To our best knowledge, Mao et al. \cite{mao2011predicting} were the first who considered data from various sources and investigated its effect on the gold prices. Rao et al. \cite{rao2013modeling} worked on similar research using opinions from twitter. Smales \cite{smales2014news}, was the first to use news sentiments from the Thompson Reuters News Analytics (TRNA) and studied its effect on the gold futures. Other works which used news sentiments in the context of commodities are \cite{feuerriegel2013news,shen2017news}. However, we believe that lack of datasets and dependency on softwares (like TRNA, RavenPack) has hindered the research focused on studying the impact of news on commodities in general and gold in particular. To the best of our knowledge, there are no publicly available datasets that can be used to create machine learning models to extract useful information from commodity news.
\par Research in the field of Natural Language Processing (NLP) has been focused on both the representation of the text as well as the classification model to process these representations. The development of context-aware pre-trained word embeddings, such as the Bidirectional Encoder Representations from Transformers (BERT) \cite{devlin2018bert}, which have been adapted to the domain of finance has shown to outperform the previous best performing models \cite{araci2019finbert}. The previously best performing algorithms have utilized deep-network architectures such as standard Recurrent Neural Network (RNN) \cite{rumelhart1986learning}, Long Short-Term Memory (LSTM) \cite{hochreiter1997long}, and Gated Recurrent Units (GRU)\cite{cho2014learning} with pretrained words embeddings such as the Global Vectors for Word Representation (GloVe) \cite{pennington2014glove,moore2017lancaster,ghosal2017iitp}. 
\par Through this paper, we introduce a human-annotated dataset of 11,412 news headlines about the gold commodity classified into nine dimensions; i.e., whether the news headline is about commodity prices, asset comparisons, or some other general information; whether the news headline is about the past price movements or future prices movements; and what is the directionality of price movements that the news suggests. In this study, we have compared the performance of various word-embeddings (Tf-Idf, GloVe and BERT) along with various machine learning algorithms; like Support Vector Machines (SVM), Recurrent Neural Networks (RNN), Long Short Term Memory (LSTM), and Gated Recurrent Units (GRU); to classify the news headlines into various dimensions. Towards the end of the paper, we perform a causality analysis, which suggests that there is a significant causal relationship between the discussions in the news and the commodity prices. In fact, an important observation that we make is that the impact of news is observable on the gold prices even 24 hours later.

\section{System Design} \label{SystemDesignSection}
\par
\begin{figure*}[ht!]
\centerline{\includegraphics[width=26pc]{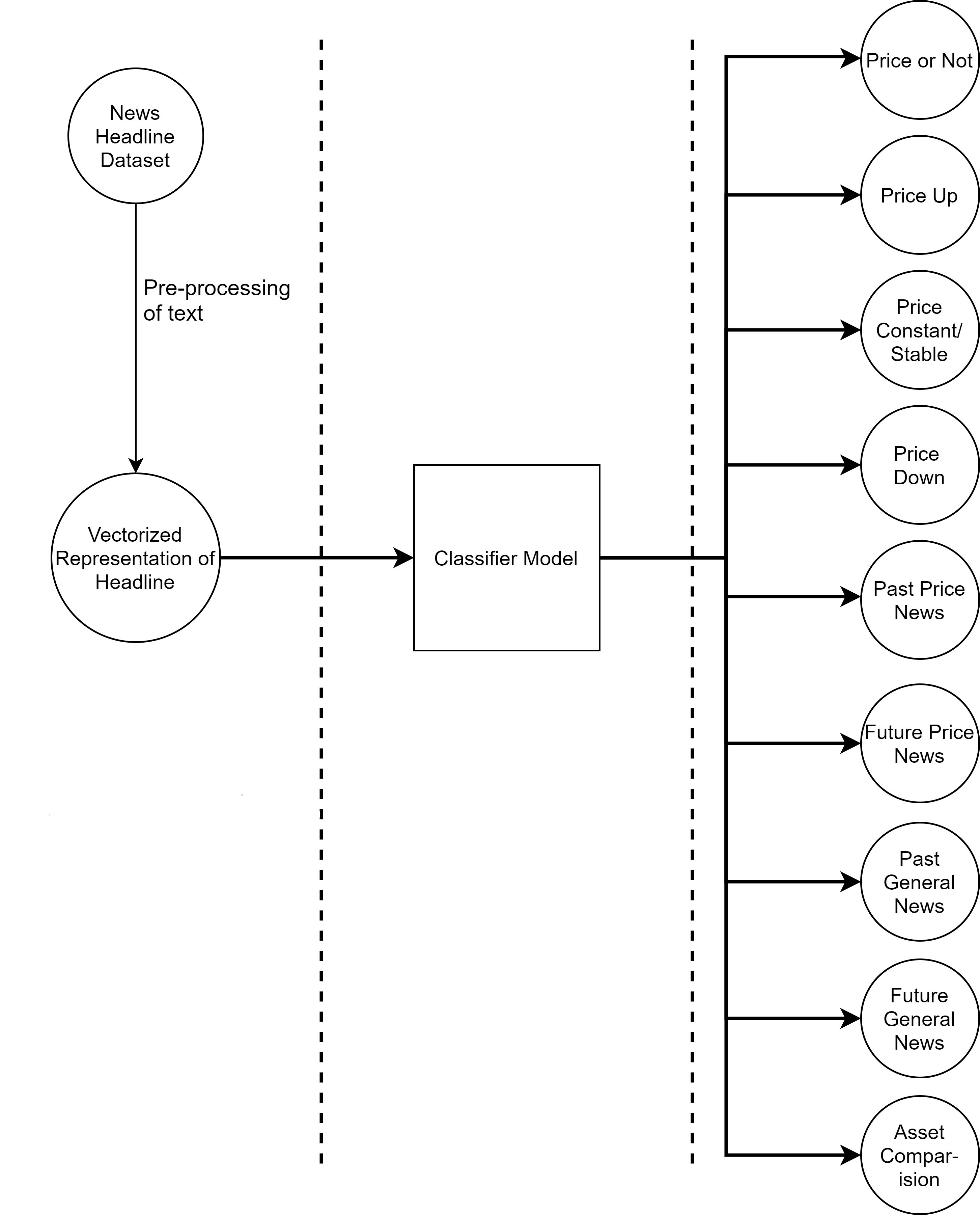}}
\caption{Overview of System Design}
\label{System Design}
\end{figure*}

 In this section, we provide the dataset creation and annotation process, followed by the learning aspects of the system \autoref{System Design} is an overview of the system that we have designed in this study. A news headline that is received in real-time is fed into our system, which first converts the headline into a vectorized representation. The vectorized representation thereafter enters a classifier model, which classifies the news into one or more of the nine classes. The classes have been identified based on our interactions with the users of the system. The most important step in the system design is the creation of the classifier model for which we needed an annotated dataset. 

\subsection{Gold News Dataset} \label{GoldNewsDataset}
With this study, we release a collection of 11,412 human-annotated news headlines dataset from around the world in the period 2000 to 2019, which are specifically about the gold commodity. This dataset was built by scraping news items from various financial news provider sites (Reuters, The Hindu, The Economic Times, Bloomberg etc.) and aggregator sites (Kitco, MetailsDaily etc.). The process of annotation required two crucial tasks; first, deciding the categories in which the news headlines must be classified into; and second, deciding the manual process of annotation.
    
\subsubsection{Categories for Annotation}
Every news item in the headline either mentions movement in prices or other general information related to the gold commodity. Hence, each of these news items were initially classified into price related news, or general news. Consider \hyperref[ex:example1]{Ex. 1} that belongs to the price category.
\begin{center}
  \ex \textit{Dec. gold settles at \$1,293.80/oz, up \$8.80, or 0.7\%.} 
  \label{ex:example1} 
\end{center}
If the news talks about price, we further classify each news headline separately on three dimensions, that tells us if the news item belongs to a particular price movement category or not. The three dimensions that talk about the price movement are:
    \begin{enumerate}[wide, labelwidth=!, labelindent=0pt]
        \item {Price Up}: This category represents the news headlines indicating if the news discusses the price heading up (irrespective of the movement being in the past or future).
        \item {Price Constant}: This category represents the news headlines indicating if the price has remained constant or stable (irrespective of the movement being in the past or future). 
        \item {Price Down}: This category represents the news headlines indicating if the news discusses the price heading down (irrespective of the movement being in the past or future). 
    \end{enumerate}
For price related news, we also look at the time period dimension, which describes the time period to which the news headline refers to with reference to the price. Consider the following news item.
    \begin{center}
        \ex \textit{Gold prices slide \$14.90, or 1.1\%, to \$1,289.80 an ounce}\label{ex:example2}
    \end{center}
    This news item talks about the decline in prices of the gold commodity in the past. Consider another news item.
    \begin{center}
        \ex \textit{Gold prices to trade higher today: Angel Commodities}\label{ex:example3}
    \end{center}
This news item mentions that the prices of gold might trade higher. Thus, in order to categorize the news events into categories based on time period, we came up with the following two categories that indicate the time period of the news item with reference to the price.
    \begin{enumerate}[wide, labelwidth=!, labelindent=0pt]
        \item Past Price Information: This category classifies the news headlines based on any past information about gold prices.
        \item Future Price Information: This category classifies the news headlines based on any future information about gold prices.
    \end{enumerate}
   
    \par News headlines can also provide general information (apart from prices) about imports, exports, production etc. Such news items were also divided based on the time period aspect, but in a different category. Consider the following news item.
    \begin{center}
        \ex\textit{Gold imports dip 8\% to \$31.72 bn in 2015-16.}\label{ex:example4}
    \end{center}
This headline tells us that gold imports have dipped. Since this news item gives us information about the gold commodity other than the price, we need to indicate that the news item talks about past information about the gold commodity, but other than the prices. Consider another news item.
    \begin{center}
        \ex\textit{WGC to form panel for setting up spot gold exchange in India.} \label{ex:example5}
    \end{center}
    This news item also does not refer to gold commodity prices, but it gives us information about an event that is going to happen in the future. Hence, we need to indicate that the news item talks about future information about the gold commodity, but other than its prices. Such future events, though not explicitly talking about gold prices, are often useful for investors and policy makers. Therefore, the two classes that talk about the time period that could be highlighted in the news headlines are:
    \begin{enumerate}[wide, labelindent=0pt]
        \item {Past General Information}: This category classifies the news headlines based on any past information other than the gold prices.
        \item {Future General Information}: This category classifies the news headlines based on any future information other than the gold prices.
    \end{enumerate}
    
    Various news headlines compare the movement of prices of two assets. We believe that capturing this information could help in gaining insights into the relationship between the gold commodity and other assets. Hence, we introduce an additional class to indicate that a news headline talks about a comparison purely in the context of the gold commodity with another asset. Consider the following news headline.
    \begin{center}
        \ex\textit{Gold notches a gain for a second day as strong dollar pauses its climb.}\label{ex:example6}
    \end{center}

The news headline in \hyperref[ex:example6]{Ex. 6} indicates that the gold commodity has gained while the dollar has paused its climb.
    
The annotations of these example headlines are shown in \autoref{tab: example_anotation}. It is important to note that a $1$ corresponding to a particular news item signifies that it belongs to that specific category, while a $0$ signifies otherwise. A news item can also belong to multiple categories (like news item in \hyperref[AnnoTabRow3]{$3^{rd}$ row} in \autoref{tab: example_anotation}).
\begin{table}[ht]
\centering
\caption{Annotation of Examples news headlines into various categories. These news headlines were taken from the dataset.}
\label{tab: example_anotation}
\resizebox{\textwidth}{!}{%
\begin{tabular}{c m{15em} C{2em} C{2em} C{2em} C{2em} C{2em} C{2em} C{2em} C{2em} C{2em}}
\toprule
\textbf{Sr. No.} &
\multicolumn{1}{c}{\begin{tabular}[c]{@{}c@{}}\textbf{News Item}\end{tabular}} & \textbf{Price or Not} & \textbf{Price Up} & \textbf{Price Const/ Stable} & \textbf{Price Down}  & \textbf{Past Price Info.} & \textbf{Future Price Info.} &\textbf{Past Gen. Info.} & \textbf{Future Gen. Info.} & \textbf{Asset Comp.}\\ \midrule
\hyperref[ex:example1]{1} & Dec. gold settles at \$1,293.80/oz, up \$8.80, or 0.7\% & 1 & 1 & 0 & 0 & 1 & 0 & 0 & 0 & 0\\ \midrule
2 & Feb. gold settles at \$1,282.30/oz, up \$5.60, or 0.4\%. & 1 & 1 & 0 & 0 & 1 & 0 & 0 & 0 & 0\\ \midrule
3 & Gold ends at a more than 1-week low, but notches slight monthly gain.\label{AnnoTabRow3}  & 1 & 1 & 0 & 1 & 1 & 0 & 0 & 0 & 0 \\ \midrule
\hyperref[ex:example2]{4} & Gold prices slide \$14.90, or 1.1\%, to \$1,289.80 an ounce & 1 & 0 & 0 & 1 & 1 & 0 & 0 & 0 & 0\\ \midrule
\hyperref[ex:example3]{5} & Gold prices to trade higher today: Angel Commodities & 1 & 1 & 0 & 0 & 0 & 1 & 0 & 0 & 0\\ \midrule 
\hyperref[ex:example4]{6} & Gold imports dip 8\% to \$31.72 bn in 2015-16. & 0 & 0 & 0 & 0 & 0 & 0 & 1 & 0 & 0\\ \midrule
\hyperref[ex:example5]{7} & WGC to form panel for setting up spot gold exchange in India & 0 & 0 & 0 & 0 & 0 & 0 & 0 & 1 & 0 \\ \midrule
\hyperref[ex:example6]{8} & Gold notches a gain for a second day as strong dollar pauses its climb. & 1 & 1 & 0 & 0 & 1 & 0 & 0 & 0 & 1 \\ \bottomrule
\end{tabular}
}
\end{table}
\subsubsection{Process of Annotation}
Our dataset was manually annotated by three human annotators who were matter experts and were given the following guidelines.
\begin{enumerate}[wide, labelindent=0pt]
    \item All annotations must be done by looking at the news headlines only. No other source like the sub-headlines,  news text, etc., must be referred to.
    \item All the annotators should independently annotate the headlines without bringing in any inherent bias that could occur from their personal views. Following the above, we arrived at three different series for the annotators. For all the cases where there was discrepancy among the annotators, a consensus-based approach was used to resolve the issue which gave us the fourth series which we refer to as the consensus series. The consensus series has been used in the paper to conduct the experiments.
\end{enumerate}

\autoref{tab: summary_category} represents the distribution of news headlines into various categories. We also report the inter-annotator's agreement score using the Cohen's Kappa statistic measure for every category. The agreement between the annotators was observed to be above 0.85 for all the categories.

\begin{table}[ht!]
\caption{This table represents the number of news from the consensus series that falls into each category (The total number of items are 11,412). We report the inter-annotator agreement using the Cohen's Kappa for all the categories.}
\centerline{
\begin{tabular}{lllll}
\toprule
\textbf{Aspects} & \textbf{Dimensions} & {\textbf{True}} & {\textbf{False}} & \textbf{\begin{tabular}[c]{@{}l@{}}Cohen's\\Kappa\end{tabular}} \\ \midrule
\multirow{5}{*}{\begin{tabular}[c]{@{}l@{}}Price\\Related\\News\end{tabular}} 
 & Price or Other & 9735 & 1677 & 0.947\\ \cmidrule{2-5}
 & Price Up & 4747 & 6665 & 0.882\\ 
   & Price Constant / Stable & 523 & 10889 & 0.895\\ 
   & Price Down & 4230 & 7182 &0.902\\ \cmidrule{2-5} 
   & Past Price Information & 9355 & 2057 & 0.912\\ 
   & Future Price Information & 381 & 11031 & 0.985 \\ \midrule
\multirow{3}{*}{\begin{tabular}[c]{@{}l@{}}Other\\News\end{tabular}} &  Other Past Information & 1598 & 9814 & 0.915\\ 
   & Other Future Information & 82 & 11330 & 0.954\\ \cmidrule{2-5} 
  & Asset Comparison & 2150 & 9262 & 0.987\\ \bottomrule
\end{tabular}%
}
\label{tab: summary_category}
\end{table}
\subsection{Classifier Models}
\par Our task is to map a text input to a binary output indicating if the news headline belongs to a particular category or not. \autoref{System Design} is a representation of the system design that we have followed in order to classify the various news headlines. After cleaning, processing and vectorizing the textual data, various classification models then take in the vectorized representations of words and give an output. Every news item is classified into various categories as shown in \autoref{System Design}. 
In our experiments we have used multiple vectorization methods as well as different classifiers. For vectorization, we used word-frequency and word-embedding based ideas, while as classifiers we have used Support Vector Machines (SVM), Recurrent Neural Networks (RNN), Long Short Term Memory (LSTM), and Gated Recurrent Units (GRU).
\par In order to compare and assess the performance of every model with each other, we define a baseline model. The baseline model uses the Term Frequency - Inverse Document Frequency (TF-IDF) weighting scheme to vectorize text and a Support Vector Machine (SVM) classifier to classify these texts into respective categories. The other models use the GloVe word-embeddings to vectorize the texts along with the other classifier algorithms and BERT approach adapted to the financial domain. The performance of every other model is compared with the baseline model. We also include the performance of pre-trained BERT approach adapted to the financial domain in our study.

\begin{figure*}[ht!]
        \centering
        \begin{subfigure}[b]{0.9\textwidth}
        \centering
        \centerline{\includegraphics[width=1.0\textwidth]{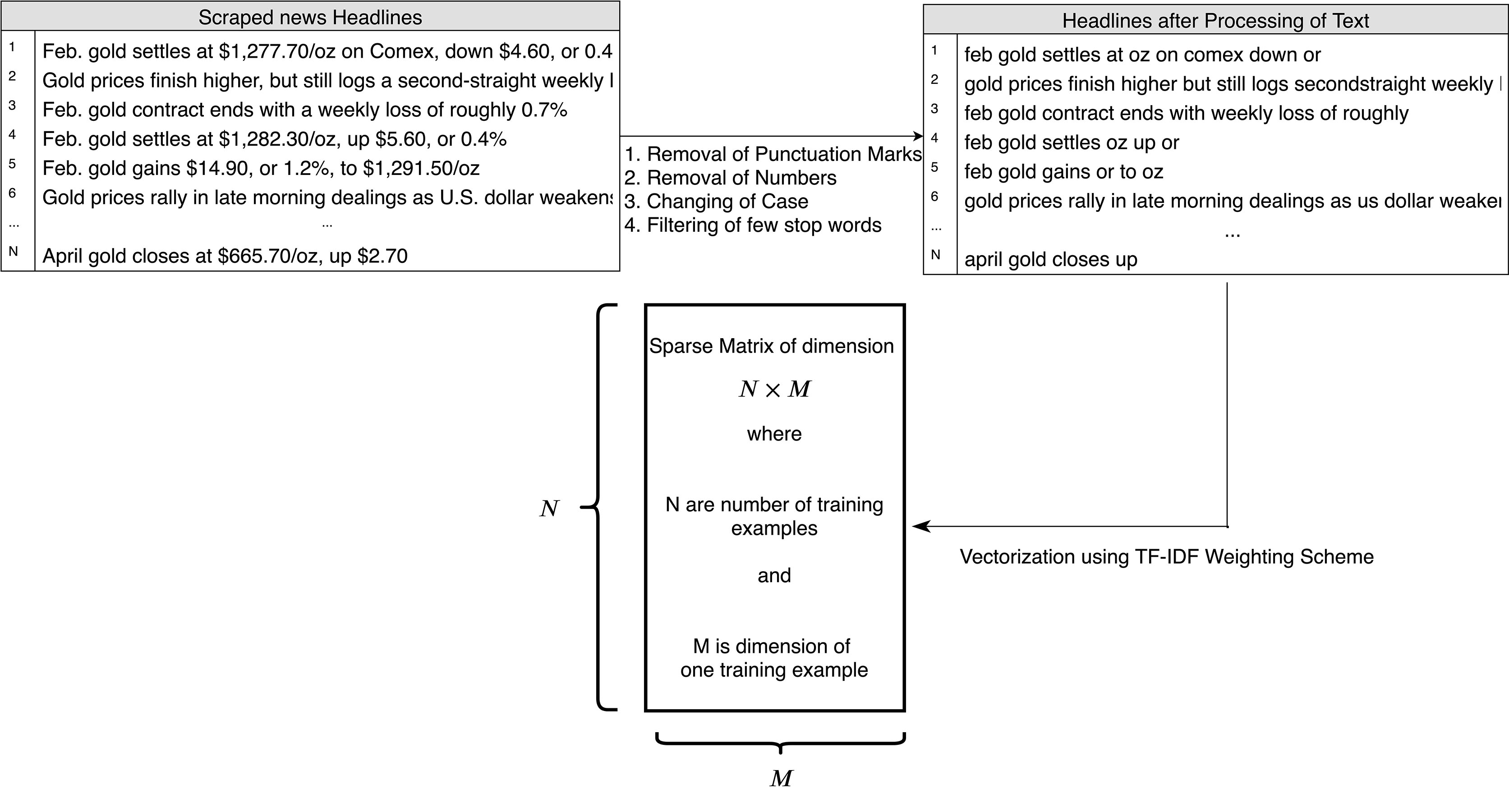}}
        \caption{Preparation of input using TF-IDF Model}
        \label{Preparation of Input for SVM}
    \end{subfigure}
    \begin{subfigure}[b]{0.9\textwidth}
        \centering
        \centerline{\includegraphics[width=1.0\textwidth]{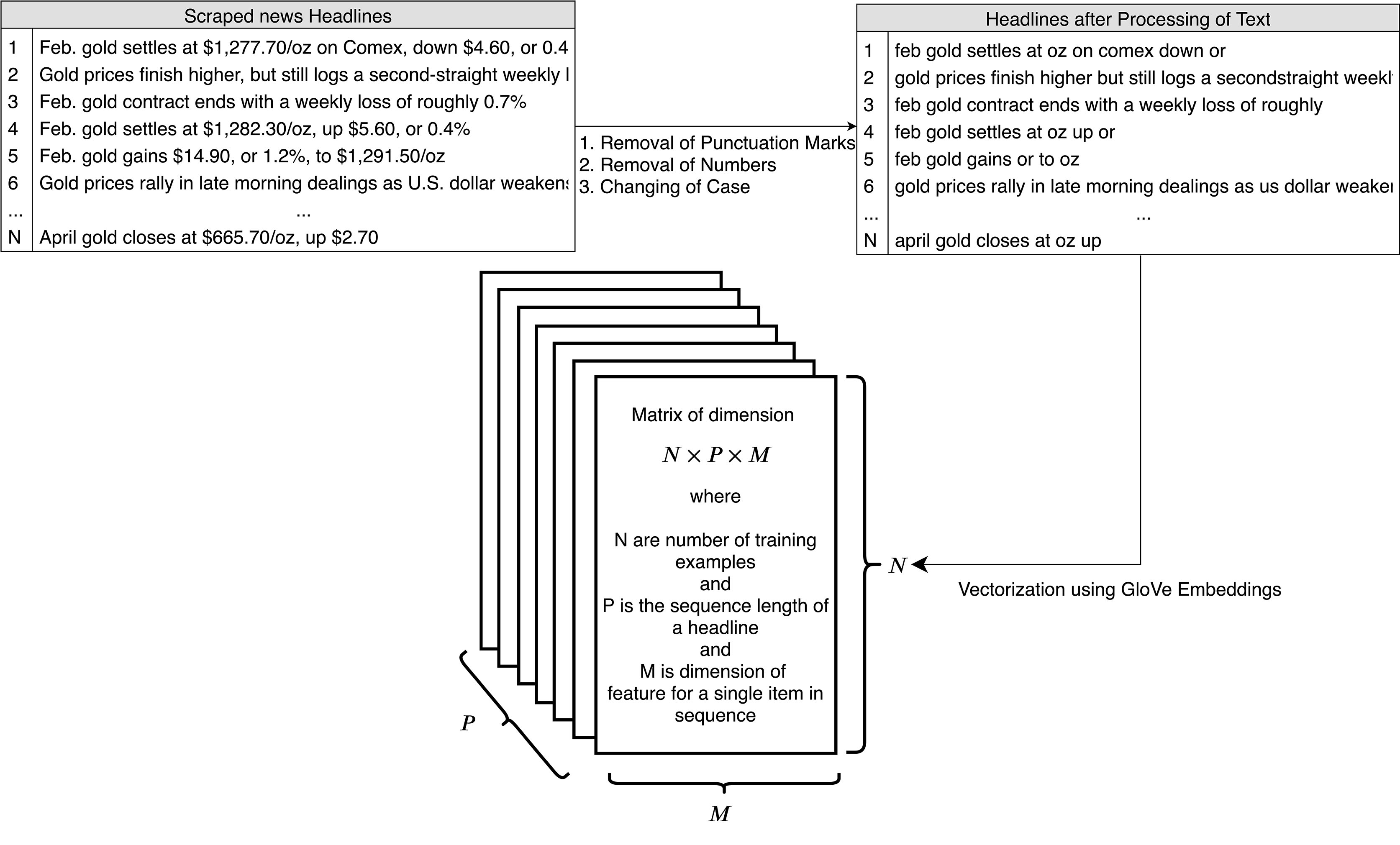}}
        \caption{Preparation of input using GloVe word-embeddings}
        \label{Preparation of Input for GloVe}
    \end{subfigure}
    \caption{Preparation of Input}
\end{figure*}

The news headline text required pre-processing before vectorization. We followed the following steps for pre-processing:
    \begin{enumerate}[wide, labelwidth=!, labelindent=0pt]
        \item {Removal of Punctuation Marks}: All kinds of punctuation mark and other special characters were removed from news headlines.
        \item {Removal of Numbers}: All numbers in the news headlines were replaced with a token ``NUM''.
        \item {Changing of Cases}: All the news headlines were converted to lower case characters.
        \item {Filtering of Stop-Words}: For the baseline model, where TF-IDF weighting scheme was used to vectorize the text, some stop words were retained while others were removed. Various stop words like \textit{up, down, above, below, under etc.}, were crucial to determine the directionality in the news headlines. Other stop words like \textit{after, before, etc.} were crucial in determining the time in the news headlines. Hence, only specific stop-words like \textit{a, an, the, of, etc.}, were filtered. For the GloVe text vectorization, no stop words were filtered.
    \end{enumerate} 
 Text vectorization is a way to represent textual data quantitatively. We have \textbf{$N$} news headlines that were scraped and pre-processed as mentioned above. Using the TF-IDF approach for the baseline model, the entire dataset of N news headlines is represented by a \textbf{$N\times M$} dimensional sparse matrix. The size of $M$ varies with the consideration of uni-gram, uni-gram---bi-gram and uni-gram---bi-gram---tri-gram tokens. This process is explained in \autoref{Preparation of Input for SVM}. 
\par The GloVe pre-trained word-embeddings are known to capture the meaning of a word through a high dimensional vector \cite{pennington2014glove}. For this research, we used the 300-dimensional vectors which were trained on 840 billion tokens through the common crawl. The outline of the process is shown in \autoref{Preparation of Input for GloVe}. It is to be noted that the entire text corpus was represented in the form of a three-dimensional matrix with size \textbf{$N\times P \times M$}.
\par To classify our dataset into various categories, we train different classifiers corresponding to each of the categories present in the dataset. For each category, the SVM classifier classified the news headline into two classes, i.e., whether it belonged to a specific category or not.
The other models that were used to compare against our baseline model are some of the more recently developed, sophisticated models. We use the GloVe algorithm to get the word-embeddings for each news headline, which was passed on to RNN, LSTM, and GRU. The Simple and bidirectional versions of these algorithms and the BERT approach adapted to financial domain led to six different classifiers apart from the baseline classifier.
\section{Results} \label{resultsSection}
In this section, we evaluate the performance of various models on categories that are related to gold prices and asset comparison. 
\autoref{results} summarizes the performance of the baseline model against 6 other models that were used to conduct the experiment. 

\begin{table*}[ht!]
\scriptsize
\centering
\caption{Precision, Recall and F1 Values on the test dataset. The percentages in brackets refer to the percentage difference between the F1 score of the baseline model and the corresponding model.}
\label{results}
\resizebox{\textwidth}{!}{%
\begin{tabular}{cccccccccc}
\toprule
\multicolumn{3}{c}{\textbf{Category}} & \textbf{Price or Not} & \textbf{Price Up} & \textbf{Price Constant} & \textbf{Price Down} & \textbf{Past Price News} & \textbf{Future Price News} & \textbf{Asset Comparison} \\ \midrule
\multicolumn{2}{c}{\multirow{3}{*}{\textbf{\begin{tabular}[c]{@{}c@{}}SVM\\ (Baseline Model)\end{tabular}}}} & \multicolumn{1}{l}{\textbf{Precision}} & 0.968 & 0.902 & 0.601 & 0.914 & \textbf{0.969} & 0.595 & 0.992 \\
\multicolumn{2}{c}{} & \multicolumn{1}{l}{\textbf{Recall}} & 0.962 & 0.947 & 0.881 & 0.951 & \textbf{0.962} & 0.949 & 0.995 \\
\multicolumn{2}{c}{} & \multicolumn{1}{l}{\textbf{F1-Score}} & 0.965 & 0.924 & 0.715 & 0.932 & \textbf{0.965} & 0.732 & 0.994 \\ \midrule
\multirow{6}{*}{\textbf{RNN}} & \multirow{3}{*}{\textbf{Simple RNN}} & \textbf{Precision} & 0.956 & 0.891 & 0.523 & 0.926 & 0.965 & 0.672 & 0.937 \\ 
 &  & \textbf{Recall} & 0.93 & 0.927 & 0.765 & 0.878 & 0.926 & 0.729 & 0.973 \\
 &  & \textbf{F1-Score} & \begin{tabular}[c]{@{}c@{}}0.943\\ (-2.283\%)\end{tabular} & \begin{tabular}[c]{@{}c@{}}0.908\\ (-0.017\%)\end{tabular} & \begin{tabular}[c]{@{}c@{}}0.622\\ (-0.13\%)\end{tabular} & \begin{tabular}[c]{@{}c@{}}0.901\\ (-0.033\%)\end{tabular} & \begin{tabular}[c]{@{}c@{}}0.945\\ (-2.117\%)\end{tabular} & \begin{tabular}[c]{@{}c@{}}0.699\\ (-4.444\%)\end{tabular} & \begin{tabular}[c]{@{}c@{}}0.955\\ (-0.039\%)\end{tabular} \\ \cmidrule(l){2-10}
 & \multirow{3}{*}{\textbf{\begin{tabular}[c]{@{}c@{}}Bidirectional \\ RNN\end{tabular}}} & \textbf{Precision} & 0.971 & 0.9 & 0.732 & 0.921 & 0.963 & 0.555 & 0.965 \\
 &  & \textbf{Recall} & 0.955 & 0.935 & 0.712 & 0.929 & 0.938 & 0.934 & 0.968 \\
 &  & \textbf{F1-Score} & \begin{tabular}[c]{@{}c@{}}0.963\\ (-0.199\%)\end{tabular} & \begin{tabular}[c]{@{}c@{}}0.917\\ (-0.758\%)\end{tabular} & \begin{tabular}[c]{@{}c@{}}0.722\\ (0.979\%)\end{tabular} & \begin{tabular}[c]{@{}c@{}}0.925\\ (-0.751\%)\end{tabular} & \begin{tabular}[c]{@{}c@{}}0.951\\ (-1.541\%)\end{tabular} & \begin{tabular}[c]{@{}c@{}}0.696\\ (-4.869\%)\end{tabular} & \begin{tabular}[c]{@{}c@{}}0.966\\ (-2.817\%)\end{tabular} \\ \midrule
\multirow{6}{*}{\textbf{LSTM}} & \multirow{3}{*}{\textbf{Simple LSTM}} & \textbf{Precision} & 0.958 & 0.917 & 0.711 & 0.913 & 0.959 & 0.656 & 0.979 \\
 &  & \textbf{Recall} & 0.971 & 0.924 & 0.757 & 0.928 & 0.949 & 0.785 & 0.989 \\
 &  & \textbf{F1-Score} & \begin{tabular}[c]{@{}c@{}}0.964\\ (-0.06\%)\end{tabular} & \begin{tabular}[c]{@{}c@{}}0.921\\ (-0.325\%)\end{tabular} & \begin{tabular}[c]{@{}c@{}}0.734\\ (2.657\%)\end{tabular} & \begin{tabular}[c]{@{}c@{}}0.921\\ (-1.18\%)\end{tabular} & \begin{tabular}[c]{@{}c@{}}0.955\\ (-1.121\%)\end{tabular} & \begin{tabular}[c]{@{}c@{}}0.715\\ (-2.298\%)\end{tabular} & \begin{tabular}[c]{@{}c@{}}0.984\\ (-1.006\%)\end{tabular} \\ \cmidrule(l){2-10}
 & \multirow{3}{*}{\textbf{\begin{tabular}[c]{@{}c@{}}Bidirectional\\ LSTM\end{tabular}}} & \textbf{Precision} & 0.96 & 0.927 & 0.698 & 0.929 & 0.964 & 0.672 & 0.958 \\
 &  & \textbf{Recall} & 0.973 & 0.921 & 0.765 & 0.917 & 0.950 & 0.835 & 0.995 \\
 &  & \textbf{F1-Score} & \begin{tabular}[c]{@{}c@{}}0.966\\ (0.154\%)\end{tabular} & \begin{tabular}[c]{@{}c@{}}0.924\\ (0\%)\end{tabular} & \begin{tabular}[c]{@{}c@{}}0.73\\ (2.098\%)\end{tabular} & \begin{tabular}[c]{@{}c@{}}0.923\\ (-0.966\%)\end{tabular} & \begin{tabular}[c]{@{}c@{}}0.957\\ (-0.836\%)\end{tabular} & \begin{tabular}[c]{@{}c@{}}0.745\\ (1.761\%)\end{tabular} & \begin{tabular}[c]{@{}c@{}}0.976\\ (-1.811\%)\end{tabular} \\ \midrule
\multirow{6}{*}{\textbf{GRU}} & \multirow{3}{*}{\textbf{Simple GRU}} & \textbf{Precision} & 0.962 & 0.909 & 0.678 & 0.925 & 0.958 & 0.672 & 0.99 \\
 &  & \textbf{Recall} & 0.972 & 0.934 & 0.789 & 0.931 & 0.964 & 0.851 & 0.995 \\
 &  & \textbf{F1-Score} & \begin{tabular}[c]{@{}c@{}}0.967\\ (0.266\%)\end{tabular} & \begin{tabular}[c]{@{}c@{}}0.921\\ (-0.325\%)\end{tabular} & \begin{tabular}[c]{@{}c@{}}0.729\\ (1.958\%)\end{tabular} & \begin{tabular}[c]{@{}c@{}}0.928\\ (-0.429\%)\end{tabular} & \begin{tabular}[c]{@{}c@{}}0.961\\ (-0.439\%)\end{tabular} & \begin{tabular}[c]{@{}c@{}}0.751\\ (2.649\%)\end{tabular} & \begin{tabular}[c]{@{}c@{}}0.993\\ (-0.101\%)\end{tabular} \\ \cmidrule(l){2-10}
 & \multirow{3}{*}{\textbf{\begin{tabular}[c]{@{}c@{}}Bidirectional\\ GRU\end{tabular}}} & \textbf{Precision} & \textbf{0.959} & 0.924 & {0.718} & 0.935 & 0.967 & 0.625 & 0.99 \\
 &  & \textbf{Recall} & \textbf{0.976} & {0.929} & {0.836} & 0.916 & 0.948 & 0.899 & 0.99 \\
 &  & \textbf{F1-Score} & \textbf{\begin{tabular}[c]{@{}c@{}}0.967\\ (0.273\%)\end{tabular}} & {\begin{tabular}[c]{@{}c@{}}0.927\\ (0.325\%)\end{tabular}} & {\begin{tabular}[c]{@{}c@{}}0.773\\ (8.112\%)\end{tabular}} & \begin{tabular}[c]{@{}c@{}}0.926\\ (-0.644\%)\end{tabular} & \begin{tabular}[c]{@{}c@{}}0.958\\ (-0.806\%)\end{tabular} & \begin{tabular}[c]{@{}c@{}}0.737\\ (0.768\%)\end{tabular} & \begin{tabular}[c]{@{}c@{}}0.99\\ (-0.402\%)\end{tabular} \\ \midrule 
 
 \multicolumn{2}{c}{\multirow{3}{*}{\textbf{\begin{tabular}[c]{@{}c@{}}BERT\end{tabular}}}} & \multicolumn{1}{l}{\textbf{Precision}} & 0.952 & \textbf{0.939} & \textbf{0.966} & \textbf{0.953} & 0.946 & \textbf{0.983} & \textbf{0.997} \\
\multicolumn{2}{c}{} & \multicolumn{1}{l}{\textbf{Recall}} & 0.95 & \textbf{0.939} & \textbf{0.942} & \textbf{0.952} & 0.944 & \textbf{0.983} & \textbf{0.996} \\&  & \textbf{F1-Score} & \begin{tabular}[c]{@{}c@{}}0.95\\ (-1.554\%)\end{tabular} & \textbf{\begin{tabular}[c]{@{}c@{}}0.939\\ (1.623\%)\end{tabular}} & \textbf{\begin{tabular}[c]{@{}c@{}}0.95\\ (32.867\%)\end{tabular}} & \textbf{\begin{tabular}[c]{@{}c@{}}0.952\\ (2.146\%)\end{tabular}} & \begin{tabular}[c]{@{}c@{}}0.945\\ (-2.117\%)\end{tabular} & \textbf{\begin{tabular}[c]{@{}c@{}}0.983\\ (34.29\%)\end{tabular}} & \textbf{\begin{tabular}[c]{@{}c@{}}0.996\\ (0.201\%)\end{tabular}} \\ \bottomrule
\end{tabular}%
}
\end{table*}

\par \autoref{results} shows that the BERT-based approach adapted to the financial domain has the best performance when compared to other models studied in this paper. Interestingly, the results show that the baseline model also works well when compared against the other models on various categories. While comparing the unidirectional models to the bidirectional models, the bidirectional performed slightly better than its counterpart. The LSTM and GRU performed better than the simple RNN, and if we compare LSTM and GRU, the latter emerges as a winner amongst the various models used. Also the time taken to train GRU was less as compared to LSTM. The categories with a huge class imbalance had a relatively low F1-Score, since we have used no technique to deal with the class imbalance. Many of these results were in line with our expectations and consistent with the literature. The BERT approach was able to better handle the categories with unbalanced classes. In general, we observe that the BERT model performs the best.

\section{Causality Analysis} \label{CausalitySection}
\par This section describes the experiment to establish the causal relationship between the news and the gold prices. We initially provide an overview of the dataset built to conduct this experiment followed by the development of metric based on the daily news items. We then go on to describe the regression model used and present our results.

\subsection{Dataset for Evaluation}
\par The task at hand was to establish a relationship between the gold news and gold prices. We establish this relationship for the gold market news and prices in the US from 2017 to 2019. The gold news items were scraped from Kitco, MetalsDaily and Reuters news sites and the prices were obtained for the same period \cite{goldPrices}. 
\par The news items from the new dataset were classified based on the direction only, i.e., if they talked about price direction up, price direction down or price direction constant. These were classified using the best model for these three categories (\autoref{results}). We build a metric called the directionality score, which evaluates the overall mood of the news in the context of price movements. 
\begin{multline*}
\text{Directionality Score } S =  \frac{N_{\text{Price Up}} - N_{\text{Price Down}}}
    {N_{\text{Price Up}} + N_{\text{Price Constant}} +  N_{\text{Price Down}}}\label{eq: directionality_score}
\end{multline*}
where $N$ is the number of news items in the respective categories.
\subsection{Model}
Using the directionality score, we build a model in order to relate the effect of news items with the gold prices. Let $S_N$ be the score on day $N$ and $P_{N}$ be the price of gold on day $N$ at 1700 hrs. The directionality scores are computed using all the news items released between 1700 hours on day $N-1$ to 1700 hours on day $N$. We study how the change in the directionality score might impact the change in gold prices as follows:
\begin{equation}
     S_{N-1} - S_{N-2} \xlongrightarrow{\makebox[1.4cm]{ $Predicts$ }} P_{N} - {P_{N-1}}\label{eq: model}
\end{equation}
For brevity, we write the \autoref{eq: model} as follows.
\begin{equation}
    S_{N-1, N-2} \xlongrightarrow{\makebox[1.4cm]{ $Predicts$ }} P_{N, N-1}
\end{equation}
Using liner regression, this can be expressed as:
\begin{equation}
    (P_{N, N-1}) = \alpha + \beta \times (S_{N-1, N-2}) + \epsilon
\end{equation}
We, therefore, set up the null hypothesis and alternate hypothesis as follows:\newline
\textbf{Null Hypothesis $(H_{O})$:} \textit{There exists no relationship between the directionality score S and price P.}\newline
\textbf{Alternate Hypothesis $(H_A)$:} \textit{There exists a relationship between the directionality score S and price P.}
\par The regression was carried out for two years separately with the first period from April 2017 to March 2018 and the second period from April 2018 to March 2019. For both these periods, we observe that $\beta$ is significant. The $p-value$ turns out to be $0.0318$ for the first period and $0.00218$ for the second period. We therefore reject the null hypothesis and conclude that there exists a causal relationship between the directionality scores $S$ and the prices $P$.
\section{Conclusion}\label{conclusionSection}
\par With this research, we have released a high-quality dataset of 11,412 news headlines about the gold commodity that has been collected from various sources around the world and annotated by human annotators on nine dimensions. This dataset can be used to analyze the various hidden meanings in the news headlines which might be of interest to investors and policy-makers. In this research, we also studied the performance of various text vectorization methods and classification algorithms in the context of our dataset. Building upon that, we performed a causality analysis, which reveals that the price related news on gold significantly impacts the prices of gold. We believe that this study will open up new avenues for news analytics research in the context of gold and also other commodities, which are known to be highly volatile in prices.
\section{Acknowledgements}
\par Ankur Sinha would like to acknowledge India Gold Policy Centre (IGPC) for supporting this study under grant number 1815012.

\end{document}